\documentclass[lettersize,journal]{IEEEtran}
\usepackage{amsmath,amsfonts}
\usepackage{algorithmic}
\usepackage{algorithm}
\usepackage{array}
\usepackage[caption=false,font=normalsize,labelfont=sf,textfont=sf]{subfig}
\usepackage{textcomp}
\usepackage{stfloats}
\usepackage{url}
\usepackage{verbatim}
\usepackage{graphicx}
\usepackage{cite}
\usepackage{booktabs} %
\usepackage{multirow} %
\usepackage{graphicx} %
\usepackage[colorlinks=true, linkcolor=blue, citecolor=blue, urlcolor=blue]{hyperref}
\usepackage[table]{xcolor}

\begin{document}

\title{Beyond Weight Adaptation: Feature-Space Domain Injection for Cross-Modal Ship Re-Identification}

\author{Tingfeng Xian*, Wenlve Zhou*, Zhiheng Zhou,~\IEEEmembership{Member,~IEEE}, Zhelin Li
\thanks{This work is Supported by the National Key Research and DevelopmentProgram of China(2022YFF0607001), National Natural Science Foundationof China (32572189), Guangdong Basic and Applied Basic Research Foundation(2023A1515010993,24202107190000687),Guangzhou City Scienceand Technology Research Projects (2023B0110011), Shaoguan Science andTechnology Research Project(230316116276286), Foshan Science and Technology Research Project(2220001018608), Zhuhai Science and TechnologyResearch Project(2320004002668). Zhongshan Science and Technology Research Project(2024A1010). (Corresponding author: Zhiheng Zhou, Zhelin Li)}
\thanks{Tingfeng Xian, Wenlve Zhou, Zhiheng Zhou are with School of Electronic and Information Engineering, South China University of Technology, Guangzhou 510641, Guangdong, China, and also with Key Laboratory of Big Data
	and Intelligent Robot, Ministry of Education, South China University of Technology, Guangzhou 510641, China (email: tingfengxian21@163.com; wenlvezhou@163.com; zhouzh@scut.edu.cn)}
\thanks{Zhelin Li is with School of  Design,South China University ofTechnology, Guangzhou 510006, China. (email: zhelinli@scut.edu.cn)}
\thanks{*Contributed to this work equally.}
}

\markboth{Journal of \LaTeX\ Class Files,~Vol.~14, No.~8, August~2021}%
{Shell \MakeLowercase{\textit{et al.}}: A Sample Article Using IEEEtran.cls for IEEE Journals}


\maketitle

\begin{abstract}
Cross-Modality Ship Re-Identification (CMS Re-ID) is critical for achieving all-day and all-weather maritime target tracking, yet it is fundamentally challenged by significant modality discrepancies. Mainstream solutions typically rely on explicit modality alignment strategies; however, this paradigm heavily depends on constructing large-scale paired datasets for pre-training. To address this, grounded in the Platonic Representation Hypothesis, we explore the potential of Vision Foundation Models (VFMs) in bridging modality gaps. Recognizing the suboptimal performance of existing generic Parameter-Efficient Fine-Tuning (PEFT) methods that operate within the weight space, particularly on limited-capacity models, we shift the optimization perspective to the feature space and propose a novel PEFT strategy termed Domain Representation Injection (DRI). Specifically, while keeping the VFM fully frozen to maximize the preservation of general knowledge, we design a lightweight, learnable Offset Encoder to extract domain-specific representations rich in modality and identity attributes from raw inputs. Guided by the contextual information of intermediate features at different layers, a Modulator adaptively transforms these representations. Subsequently, they are injected into the intermediate layers via additive fusion, dynamically reshaping the feature distribution to adapt to the downstream task without altering the VFM's pre-trained weights. Extensive experimental results demonstrate the superiority of our method, achieving State-of-the-Art (SOTA) performance with minimal trainable parameters. For instance, on the HOSS-ReID dataset, we attain 57.9\% and 60.5\% mAP using only 1.54M and 7.05M parameters, respectively. The code is available at https://github.com/TingfengXian/DRI.
\end{abstract}
%

\section{Introduction}

\par
\begin{figure}[!t]
	\centering
	\includegraphics[width=0.48\textwidth]{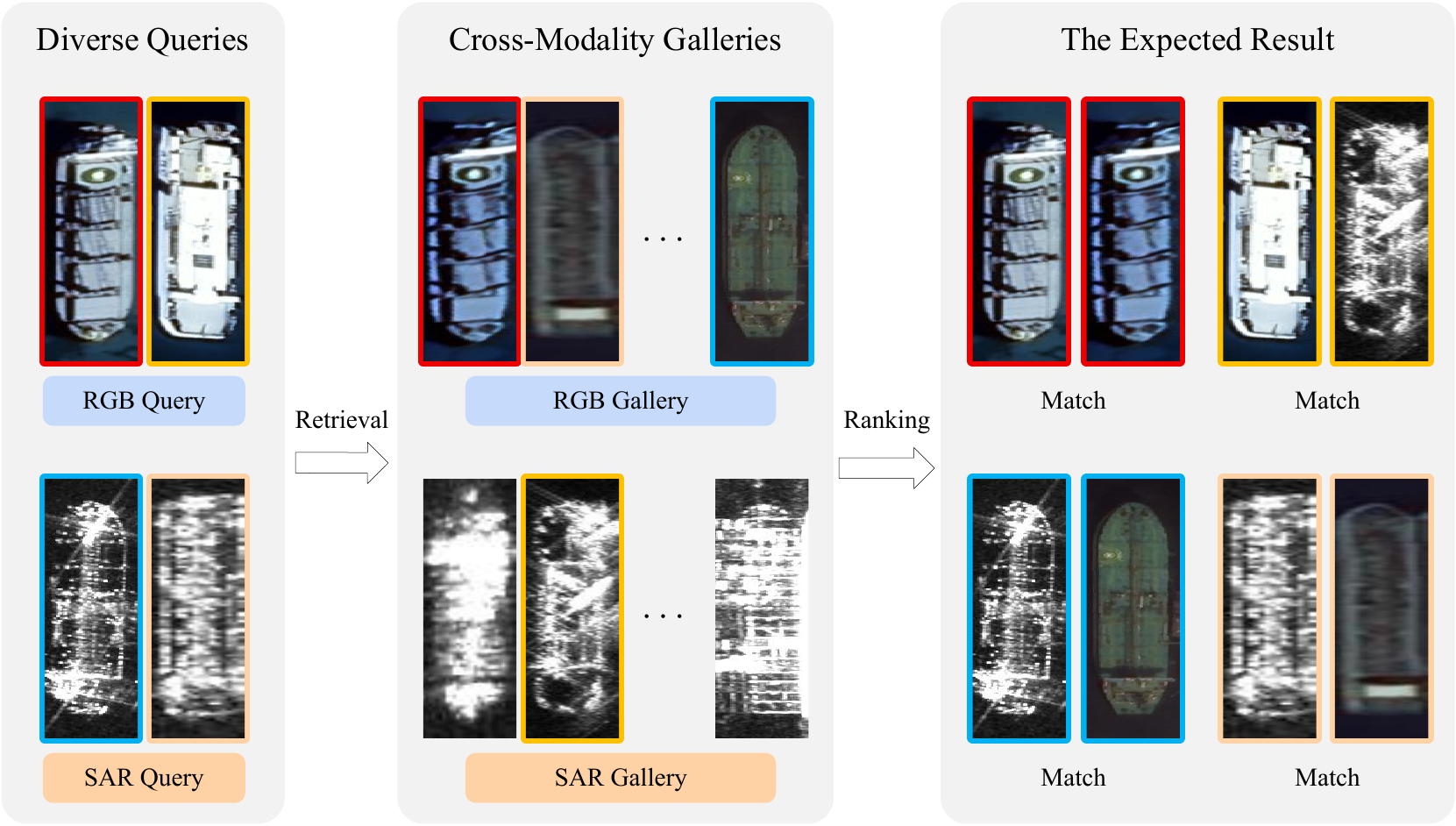} 
	\caption{\textbf{Schematic illustration of the Cross-Modality Ship Re-Identification task.} 
		Take optical-SAR Ship Re-ID as an example, the process is divided into three stages, frames of the same color indicate that the ids of Ships are the same. }
	\label{fig:CMS-ReID}
\end{figure}

 Cross-Modality Ship Re-Identification (CMS Re-ID) aims to retrieve specific ship targets from massive and heterogeneous remote sensing images \cite{wang2025cross}. Fig.\ref{fig:CMS-ReID} provides a schematic illustration of CMS Re-ID. As a challenging task in the field of remote sensing Earth observation, CMS Re-ID is crucial for achieving all-day and all-weather maritime target tracking, holding extensive application value in maritime security, search and rescue, and shipping monitoring and analysis \cite{yang2021enhanced, wu2023ristrack, kong2024ts}. Constrained by imaging mechanisms, single-modality data often falls short of meeting all-day and all-weather monitoring requirements. Addressing the weather dependency of optical images and the scale limitations of SAR data, Literature \cite{wang2025cross} highlighted the necessity of multi-modal data and proposed an Optical-SAR CMS Re-ID dataset named HOSS-ReID to bridge this gap. Meanwhile, with the advancement of drone remote sensing technology, Literature \cite{xu2025cmshipreid} constructed the multi-source heterogeneous CMShipReID dataset by integrating Optical, Near-Infrared (NIR), and Thermal Infrared (TIR) sensors, providing richer data support and research perspectives for the field.

\par The core challenge of CMS Re-ID lies in the significant discrepancy between heterogeneous modality images. Mainstream solutions typically rely on explicit modality alignment strategies to align feature spaces \cite{zhao2025freefusion, jiang2025harmonized}. However, this paradigm depends on constructing large-scale paired datasets. The Platonic Representation Hypothesis postulates that neural networks, trained with disparate objectives and data modalities, are converging towards a shared statistical model of reality in their representation spaces, with this convergence intensifying as model capabilities improve \cite{huh2024position}. This hypothesis has been further corroborated by recent studies. Specifically, \cite{han2025learning} demonstrated that Large Language Models (LLMs) trained exclusively on text can acquire prior capabilities transferable to visual tasks without ever seeing images. Furthermore, \cite{kaushik2025universal} conducted a large-scale analysis of over 1,100 models, revealing a striking phenomenon: the parameters of networks across different tasks, datasets, and architectures tend to converge into a common, low-dimensional universal subspace. Inspired by these findings, we explore a critical question: Can Vision Foundation Models (VFMs) be leveraged to bridge the modality gap in CMS-ReID? Given that VFMs are pre-trained on massive-scale datasets, the hypothesis suggests they exhibit the strongest convergence towards the underlying reality. We posit that although the visual manifestations of the same ship identity diverge significantly across different modalities, they intrinsically correspond to the same underlying reality. Therefore, by exploiting the superior representational convergence of VFMs, we aim to effectively surmount the modality disparities inherent in CMS-ReID.

\begin{figure}[!t]
	\centering
	\includegraphics[width=0.45\textwidth]{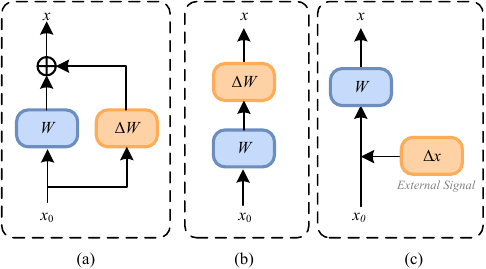}
	\caption{\textbf{Comparison between Weight-based PEFT and Feature-based PEFT.} (a) Reparameterization-based PEFT (e.g., LoRA) reconstructs the original weight matrix $W$ of the VFM by introducing a low-rank incremental matrix $\Delta W$. (b) Additive PEFT (e.g., Adapter) introduces new weight matrices and integrates them into the original network, thereby modifying the model's topology. Both (a) and (b) operate within the weight space. (c) Our proposed Feature-based PEFT introduces external signals to explicitly reshape the feature distribution. This approach maximizes the preservation of general knowledge without altering the weight structure of the VFM.}
	\label{overview}
\end{figure}

\par Existing VFMs adaptation strategies often employ Parameter-Efficient Fine-Tuning (PEFT) \cite{ding2023parameter} to avoid the prohibitive costs of full-parameter fine-tuning. However, directly applying generic PEFT methods yields suboptimal results in cross-modality ship Re-ID. From an optimization perspective, such methods typically formulate fine-tuning as learning an incremental matrix $\Delta W$ superimposed on the frozen weights (e.g., LoRA \cite{hu2022lora} assumes $\Delta W$ is a low-rank matrix). However, when there is a massive domain distribution gap between the VFMs and the downstream task, the intrinsic dimension theory suggests that the model requires a higher-rank update to achieve effective alignment. A higher rank implies a demand for more data and computational power. Furthermore, Literature \cite{zhou2024unsupervised} points out that during the adaptation of VFMs to downstream tasks, only a negligible portion of weight adjustments is beneficial. Consequently, blind fitting within the full weight space is often inefficient due to the inclusion of substantial redundant updates.

\par Based on these insights, we propose shifting the optimization perspective from the weight space to the feature space. Rather than struggling to find effective weight offsets within the weight space, it is more effective to perform explicit calibration directly on intermediate features. This approach completely preserves the efficient pre-trained weights, preventing the model from forgetting prior knowledge during fine-tuning. We utilize a lightweight, learnable encoder to extract domain representations from raw data and inject them into the intermediate features. This reshapes the feature distribution to adapt to the downstream task without altering the VFMs weights. As illustrated in Fig.\ref{overview}, existing PEFT methods generally fall into two categories: (a) Reparameterization-based methods (e.g., LoRA), which reconstruct original weights via low-rank matrices; and (b) Additive methods (e.g., Adapter \cite{houlsby2019parameter}), which modify the model architecture by inserting additional layers. Both paradigms essentially operate within the weight space. In contrast, as shown in Fig.\ref{overview}(c), our method shifts the optimization perspective to the feature space. By abandoning adjustments in the weight space and focusing on directly calibrating intermediate features, our approach maximizes the preservation of the generalizable knowledge inherent in the pre-trained model.

\par Specifically, this paper proposes a fine-tuning strategy based on Domain Representation Injection (DRI). Targeting a frozen VFM, we design a lightweight Offset Encoder (OE) aimed at mining domain-specific representations rich in modality and identity attributes from raw inputs. These representations are modulated by Modulator (Mod) and then superimposed onto the intermediate feature layers of the VFM, thereby reshaping its distribution in the feature space to better suit the specific domain re-identification task. Notably, this module features a significant plug-and-play capability and requires updating only a minimal number of parameters, achieving negligible training overhead and efficient deployment. The main contributions of this paper are summarized as follows:

\begin{itemize}
	\item[(1)] Diverging from traditional PEFT paradigms that adapt models by optimizing the weight space, we explore the potential of feature-space fine-tuning. We propose a plug-and-play Domain Representation Injection strategy that directly calibrates the VFM distribution by externally introducing domain representations.
	\item[(2)] We propose a lightweight injection framework comprising an OE for domain extraction and a Mod for adaptive feature transformation. This synergistic design facilitates effective domain injection via an additive fusion mechanism with minimal computational overhead.
	\item[(3)] Extensive experiments demonstrate the superiority of the proposed method. On two benchmark datasets, our method achieves State-of-the-Art (SOTA) performance with only 1.54M/6.78M trainable parameters. On HOSS-ReID, we achieve 57.9\% and 60.5\% mAP with 1.54M and 7.05M parameters, respectively. Furthermore, on CMShipReID, our method sets new records with 6.78M parameters, achieving 84.1\%/84.9\% mAP for TIR-VIS, 72.6\%/71.8\% mAP for NIR-VIS, and 67.5\%/64.3\% mAP for NIR-TIR scenarios.
\end{itemize}

\section{Related Work}

\subsection{Cross-Modality Ship Re-Identification}

\par Constrained by the difficulty of acquiring multi-modal remote sensing data, early research progress in CMS Re-ID was relatively slow. However, with the rapid advancement of remote sensing technology, the availability of multi-source heterogeneous data has significantly improved \cite{zhao2024hvt, he2025dogan, liu2025aerial}. Consequently, ship Re-ID has expanded beyond single-modality limitations, evolving towards multi-modal synergy involving Optical, SAR, and Infrared data.

\par Regarding the Optical-SAR cross-modality task, Literature \cite{wang2025cross} released the first dataset, HOSS ReID, and highlighted that this task presents greater challenges than Optical-IR Re-ID due to the substantial disparity in imaging mechanisms. To address this, the authors proposed TransOSS, a method utilizing a two-stage strategy comprising "CLIP-like contrastive pre-training followed by fine-tuning" based on the Transformer architecture. For Optical-IR cross-modality tasks, Literature \cite{zhang2024third} proposed a Third-Modality Collaborative Learning Approach to mitigate modality discrepancies by generating an auxiliary modality to facilitate alignment. Meanwhile, Literature \cite{wen2024cross} introduced a method based on Deep Alignment Decomposition, which employs geometric and semantic alignment to alleviate global feature biases and captures fine-grained features via adaptive local decomposition. Furthermore, Literature \cite{xu2025cmshipreid} constructed a UAV-based Optical-IR dataset, CMShipReID, covering visible, near-infrared, and thermal infrared modalities.

\par Although existing studies have confirmed that modality discrepancy is the core challenge in CMS Re-ID, the potential of VFMs in addressing this issue remains underexplored. Consequently, this paper proposes a feature-space PEFT method, aiming to efficiently transfer the high generalization capability of VFMs to CMS Re-ID tasks.

\subsection{Parameter-Efficient Fine-Tuning}

\par As deep learning transitions into the era of VFMs, the traditional paradigm of "pre-training followed by full fine-tuning" faces severe challenges due to prohibitive computational costs and storage overheads. Consequently, PEFT has emerged as a pivotal technique, aiming to adapt general-purpose large models to specific downstream tasks by updating only a small subset of parameters or introducing minimal external modules.

\par Existing PEFT strategies can be broadly categorized into addition-based and reparameterization-based methods. Additive methods, represented by Adapters \cite{houlsby2019parameter, chen2022vision, yin20255}, typically embed lightweight bottleneck modules serially or in parallel within pre-trained Transformer layers, utilizing non-linear projections to adapt intermediate features. Conversely, reparameterization methods, epitomized by LoRA \cite{hu2022lora}, rely on the Intrinsic Dimension hypothesis. They freeze the pre-trained weights and inject trainable rank-decomposition matrices into linear layers, approximating weight updates via low-rank approximations. Additionally, prompt-based approaches like Visual Prompt Tuning (VPT) \cite{jia2022visual} introduce learnable tokens into the input sequence to steer the model's attention. Despite their diverse implementations, these methods share a common philosophy: achieving adaptation by implicitly modifying the model's behavior within the weight optimization space.

\par While PEFT provides a feasible pathway for applying VFMs to CMS Re-ID, existing methods are primarily confined to this weight-space optimization. However, the substantial distribution shift among different modalities implies a high-dimensional transformation, often necessitating a high-rank weight update to bridge the discrepancy. This fundamentally contradicts the "low-rank update" assumption underlying methods like LoRA, making it difficult to achieve effective alignment in cross-modal scenarios with limited data. In light of this, we shift the optimization perspective from the weight space to the feature space. By directly calibrating intermediate representations and maximizing the preservation of the VFMs' general knowledge without altering the original weight matrices, we achieve efficient and robust adaptation for the CMS Re-ID task.

\subsection{Modulation in Computer Vision}

\par Feature modulation plays a pivotal role in computer vision, where external signals are injected as conditions to dynamically alter the statistical properties of neural network features. This paradigm first gained prominence in style transfer tasks. Specifically, Adaptive Instance Normalization (AdaIN) \cite{huang2017arbitrary} demonstrated that style could be effectively transferred by aligning feature mean and variance. Subsequently, this concept was generalized into Feature-wise Linear Modulation (FiLM) \cite{perez2018film}, which applies affine transformations (scaling and shifting) to feature maps conditioned on external inputs, facilitating effective multi-modal fusion.

\par In the evolution of diffusion models, modulation mechanisms have served as a cornerstone for conditional generation. In DDPM \cite{ho2020denoising} and Guided Diffusion based on the U-Net architecture \cite{dhariwal2021diffusion}, the timestep $t$ is injected by modulating the normalization layers of residual blocks to control the denoising process. More recently, adapting to the shift towards Transformer architectures, the Diffusion Transformer (DiT) \cite{peebles2023scalable} was introduced with the adaLN-Zero mechanism. Drawing inspiration from StyleGAN \cite{karras2019style}, adaLN-Zero injects timesteps $t$ and class labels $c$ into ViT via adaptive layer normalization with zero initialization, enabling powerful conditional generation with minimal computational overhead.

\par Inspired by these advancements—particularly the efficacy of adaLN in modulating Transformer features—we propose a novel feature-based PEFT strategy for CMS Re-ID. Unlike the scalar conditions used in DiT, raw images containing rich ID and modality-specific semantics are leveraged as external signals. By modulating the feature distribution of the VFM conditioned on these signals, the proposed method effectively steers the pre-trained model to bridge the significant modality gap without the need for adaption of weight space.

\section{Methods}

\begin{figure*}[!t]
	\centering
	\includegraphics[width=0.90\textwidth]{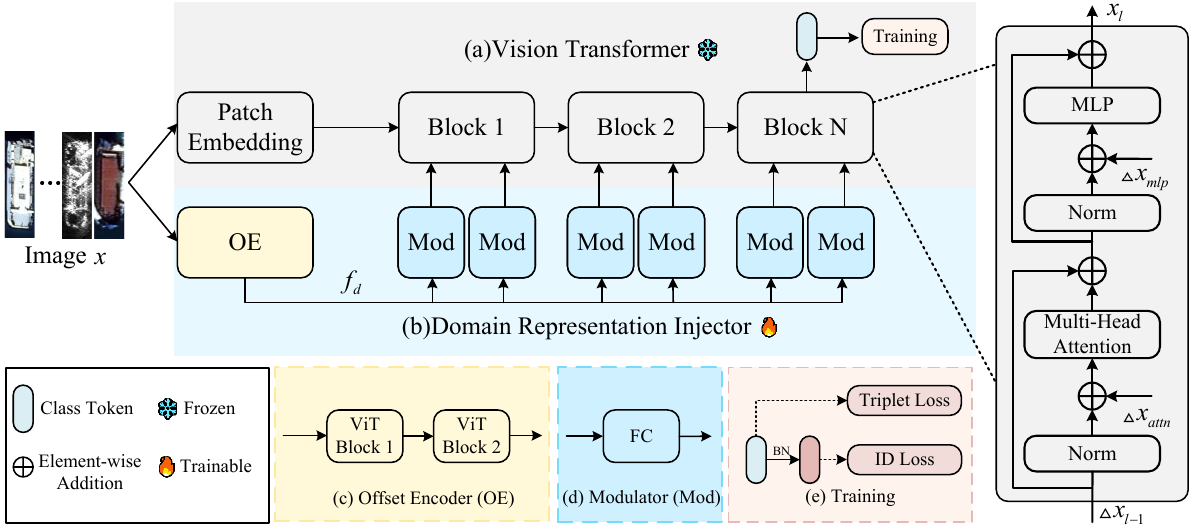}
	\caption{\textbf{Overview of the proposed Domain Representation Injection (DRI) framework.} The architecture comprises two primary components: a fully frozen VFM (initialized with DINOv3) and a trainable Domain Representation Injector. Given an input image, the Offset Encoder (OE) extracts the domain-specific representation $f_d$. Subsequently, Modulators (Mod) adaptively transform $f_d$ into specific feature deviations $\Delta x$ based on the context of different layers. These deviations are injected into the Attention and MLP layers of the VFM via additive fusion to reshape the feature distribution. The model is trained using a combination of Triplet Loss and ID Loss. }
	\label{framework}
\end{figure*}

\par As illustrated in Fig. \ref{framework}, the proposed framework consists of two core components: a fully frozen VFM and a trainable Domain Representation Injector. Leveraging its superior performance and powerful generalizable representations, we adopt the ViT-based DINOv3 \cite{simeoni2025dinov3} as the VFM. The Domain Representation Injector comprises an OE and a Mod, where the OE is instantiated as a lightweight two-layer Vision Transformer (ViT) \cite{dosovitskiy2020image} and the Mod is implemented as a simple linear layer. We extract the first token (i.e., the [CLS] token) from the final output of the VFM to serve as the global representation $f_g$.

\par During the training phase, we follow the training recipes in \cite{wang2025cross}: the Triplet Loss is computed directly using $f_g$, while the ID Loss (Cross-Entropy Loss) is calculated employing the BNNeck strategy \cite{luo2019bag}. Consequently, the overall objective function is formulated as:
\begin{equation}
	\mathcal{L}_{total} = \mathcal{L}_{Triplet} + \mathcal{L}_{ID}
	\end{equation}
\par In the inference phase, we employ Euclidean distance to measure the similarity between the global representations $f_g$ of the query and gallery samples. Based on the computed distance matrix, retrieval ranking is performed to achieve cross-modal object re-identification.

\subsection{Preliminaries}
\par Existing generic PEFT methods primarily focus on parameter-efficient fine-tuning within the weight space. Taking the representative LoRA \cite{hu2022lora} as a prime example, inspired by the intrinsic dimension hypothesis, it postulates that the weight updates $\Delta W$ of pre-trained models during adaptation are intrinsically subject to a low-rank constraint. Formally, given a frozen pre-trained weight matrix $W_0 \in \mathbb{R}^{d \times k}$, LoRA decomposes $\Delta W$ into the product of two low-rank matrices, $B \in \mathbb{R}^{d \times r}$ and $A \in \mathbb{R}^{r \times k}$, where the rank satisfies $r \ll \min(d, k)$. Letting $x$ denote the input, the forward propagation process of LoRA is modeled as:
\begin{equation}
	h = W_0 x + \Delta W x = W_0 x + B A x
\end{equation}
\par From the perspective of weight optimization, this is equivalent to re-parameterizing the fine-tuned weights as follows:
\begin{equation}
	W = W_0 + \Delta W = W_0 + B A
\end{equation}
\par The rank $r$ determines the upper bound of the representation capability and the fitting complexity of the VFM during fine-tuning. Generally, a larger domain distribution gap between the VFM and the downstream task necessitates a higher rank to bridge this discrepancy, which inevitably increases the demand for training data and computational resources. However, the availability of sufficient training data is not always guaranteed. Specifically, in the Optical-SAR Ship Re-ID scenario, the extreme paucity of samples constitutes a major bottleneck \cite{wang2025cross}. This renders fine-tuning strategies based on large-scale or high-rank parameters ill-suited for effective convergence, thereby severely limiting the performance ceiling of the model. Furthermore, existing study \cite{zhou2024unsupervised} indicate that only a negligible portion of weight adjustments during fine-tuning contributes to downstream tasks. Consequently, blind fitting within the full weight space is often inefficient due to the inclusion of substantial redundant updates.

\subsection{Offset Encoder}

\par Given the dual dilemmas of data scarcity and computational efficiency associated with direct optimization in the weight space, we shift the optimization perspective from the weight space to the feature space. As illustrated in Fig. \ref{overview}. Specifically, in contrast to the paradigm in Eq. (2) that adjusts weights by introducing a parameter update $\Delta W$, we keep the pre-trained weights $W_0$ frozen. Instead, we reshape the input by injecting a feature deviation $\Delta x$. Consequently, the forward propagation process of the model is reformulated as:
\begin{equation}
	h = W_0 x + W_0 \Delta x = W_0 (x + \Delta x)
\end{equation}
\par From the perspective of feature optimization, this operation is equivalent to reshaping the original input feature $x_0$ via an additive deviation $\Delta x$:
\begin{equation}
	x = x_0 + \Delta x
\end{equation}
\par
\begin{figure}[!t]
	\centering
	\includegraphics[width=0.35\textwidth]{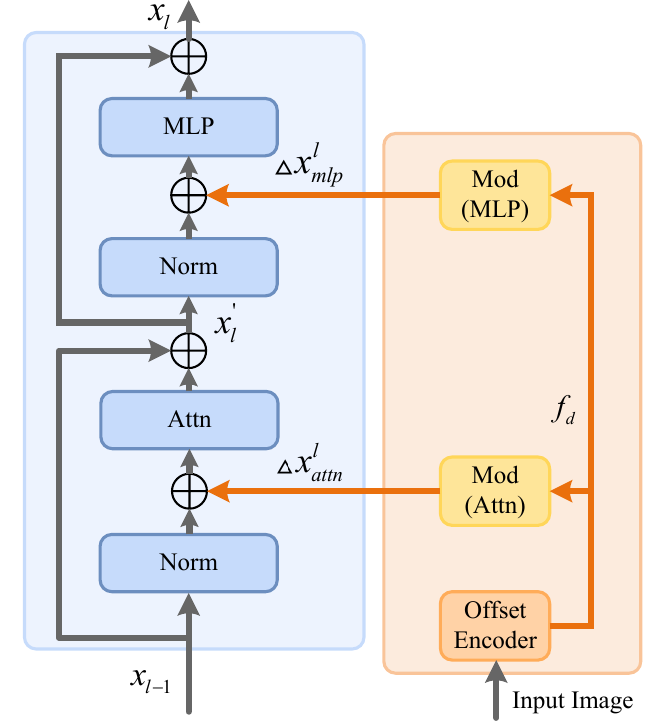}
	\caption{\textbf{Illustration of the proposed Domain Representation Injector.} For the $l$-th block of the VFM with input feature $\mathbf{x}_{l-1}$, the domain representation $f_d$ is first derived via the Offset Encoder. Subsequently, $f_d$ is processed by the modulators $\text{Mod}_{\text{attn}}$ and $\text{Mod}_{\text{mlp}}$ to generate the specific feature deviations $\Delta \mathbf{x}^{\ell}_{\text{attn}}$ and $\Delta \mathbf{x}^{\ell}_{\text{mlp}}$, respectively. Finally, $\Delta \mathbf{x}^{\ell}_{\text{attn}}$ is injected into $\mathbf{x}_{l-1}$ via additive fusion, while $\Delta \mathbf{x}^{\ell}_{\text{mlp}}$ is added to the intermediate feature $\mathbf{x}'_{l}$.}
	\label{DRI}
\end{figure}

 Therefore, the core problem transforms into how to effectively model this feature deviation $\Delta x$. To this end, we design a lightweight and learnable OE, which aims to capture domain representations rich in modality and identity information from the raw input image. As illustrated in Fig. \ref{DRI}. Specifically, the OE is instantiated as a two-layer ViT incorporating Rotary Position Embeddings (RoPE) \cite{su2024roformer}. Given an input image $\mathbf{x}$, the domain representation $f_d$ is computed as follows:
\begin{align}
	\mathbf{z}_0 &= [\mathbf{x}_{\text{domain}}; \mathbf{x}_p^1 \mathbf{E}; \cdots ; \mathbf{x}_p^N \mathbf{E}], & \mathbf{E} \in \mathbb{R}^{(P^2 \cdot C) \times D} \\
	\mathbf{z}'_\ell &= \text{Attn}(\text{Norm}(\mathbf{z}_{\ell-1})) + \mathbf{z}_{\ell-1}, & \ell = 1, 2,...,L \\
	\mathbf{z}_\ell &= \text{MLP}(\text{Norm}(\mathbf{z}'_\ell)) + \mathbf{z}'_\ell, & \ell = 1, 2,...,L \\
	f_d &= \text{Norm}(\mathbf{z}_L^0) &
\end{align}

\par where $\mathbf{z}_0$ denotes the input embedding sequence. $\mathbf{x}_{\text{domain}}$ represents a learnable domain token initialized to capture global domain information, similar to the [CLS] token in standard ViT. $\mathbf{x}_p^i$ refers to the $i$-th flattened patch of the input image, and $\mathbf{E} \in \mathbb{R}^{(P^2 \cdot C) \times D}$ is the linear projection matrix mapping patches to the embedding dimension $D$. $N$ is the total number of patches. $\ell$ denotes the layer index, with L=2 indicating the two-layer structure of the OE. $\text{Norm}$, $\text{Attn}$, and $\text{MLP}$ denote Layer Normalization, Multi-Head Self-Attention, and Multi-Layer Perceptron, respectively.

\subsection{Domain Representation Modulation}

\par Although the Offset Encoder extracts a compact global domain representation $f_d$, directly injecting this identical vector into all layers of the VFM is suboptimal. Since the feature distributions evolve layer-by-layer throughout the forward propagation of the VFM, each block possesses a distinct feature context and thus requires a specific calibration signal. Furthermore, within each Transformer block, the Multi-Head Self-Attention (MHSA) module focuses on modeling global spatial dependencies, whereas the Multi-Layer Perceptron (MLP) module is responsible for channel-wise feature transformation and non-linearity.

\par To address these layer-specific and module-specific requirements, we introduce a Layer-Adaptive Modulation mechanism. As illustrated in Fig. \ref{DRI}, rather than utilizing a fixed global deviation, we employ parameter-independent modulators for the $\ell$-th block to adaptively project $f_d$ into the corresponding feature space. This process generates precise, context-aware feature deviation terms $\Delta \mathbf{x}$:
\begin{equation}
	\Delta \mathbf{x}^{\ell}_{\text{attn}} = \text{Mod}^{\ell}_{\text{attn}}(f_{d}) = \mathbf{W}^{\ell}_{\text{attn}} f_d + \mathbf{b}^{\ell}_{\text{attn}}
	\end{equation}
\begin{equation}
	\Delta \mathbf{x}^{\ell}_{\text{mlp}} = \text{Mod}^{\ell}_{\text{mlp}}(f_{d}) = \mathbf{W}^{\ell}_{\text{mlp}} f_d + \mathbf{b}^{\ell}_{\text{mlp}}
\end{equation}

\par where $\mathbf{W}_{*}^{\ell}$ and $\mathbf{b}_{*}^{\ell}$ denote the learnable weights and biases corresponding to each module (i.e., $* \in \{\text{attn}, \text{mlp}\}$), initialized to zero to ensure a stable start from the pre-trained distribution.

\par Subsequently, these modulated representations are injected into the VFM via additive fusion. Crucially, we perform this injection after Layer Normalization. This design ensures that the learned domain deviations act as a direct calibration signal on the standardized features, effectively guiding the subsequent Attn/MLP computations without being suppressed by normalization statistics. Formally, for the $\ell$-th block with input $\mathbf{x}_{\ell-1}$, the forward propagation is formulated as:
\begin{align}
	\mathbf{x}'_{\ell} &= \mathbf{x}_{\ell-1} 
	+ \text{Attn}(\text{Norm}(\mathbf{x}_{\ell-1}) + \Delta \mathbf{x}^{\ell}_{\text{attn}}) \\
	\mathbf{x}_{\ell} &= \ \mathbf{x}'_{\ell} \ 
	+ \text{MLP}(\text{Norm}(\mathbf{x}'_{\ell}) + \Delta \mathbf{x}^{\ell}_{\text{mlp}})
\end{align}

\par By explicitly decoupling the modulation for each layer and module, our method ensures that the domain adaptation is fine-grained and adaptive to the evolving feature contexts, maximizing the retention of the VFM's general knowledge while efficiently bridging the modality gap.

\section{Experiments}
\subsection{Datasets}

To validate the effectiveness of our proposed method, we conducted extensive experiments on two mainstream public cross-modal ship re-identification datasets: HOSS-ReID \cite{wang2025cross} and CMShipReID \cite{xu2025cmshipreid}. The details of these two datasets are described below:

\begin{itemize}
\item \textbf{HOSS-ReID:} This Optical-SAR dataset contains 1,832 images across 449 ship IDs (1,065 optical, 767 SAR), plus 163 distractor IDs labeled as "-1". Notably, it provides object size and aspect ratio for all ships. Following the original protocol, the training set includes 361 IDs with 1,063 images (574 optical, 489 SAR). The test set comprises a query of 176 images from 88 IDs (split evenly between modalities) and a gallery containing the entire 1,832-image dataset (targets and distractors).

\item \textbf{CMShipReID:} This Visible(VIS)-Infrared(IR) dataset subdivides IR into Near-Infrared (NIR) and Thermal-Infrared (TIR), comprising 8,337 images of 138 IDs. It includes 3,209 VIS images (138 IDs), 2,668 NIR images (134 IDs), and 2,460 TIR images (138 IDs). Following the standard partition, 100 IDs are used for training. The test set contains the remaining IDs: 38 for the VIS-TIR task and 34 for the VIS-NIR and TIR-NIR task.
\end{itemize}

\subsection{Implementation Details}

\par For the HOSS-ReID dataset, we followed the data augmentation strategies consistent with the original paper, including random horizontal flipping, cropping, and random erasing. The input images were resized to $128 \times 256$. Furthermore, following the implementation protocol in the original paper, we encoded the ship size and aspect ratio provided by the dataset into a token via a linear layer. Subsequently, this token was concatenated after the [CLS] token of the VFM. 
\par For the CMShipReID dataset, consistent with its original work, we did not employ any data augmentation, and the images were resized to $224 \times 224$. For both datasets, we employed the Stochastic Gradient Descent (SGD) optimizer with an initial learning rate set to 0.0015 and a batch size of 32. The models were trained for a total of 200 epochs. 
\par All experiments were conducted on a single NVIDIA RTX 4090 GPU equipped with 24GB of RAM, using the PyTorch framework. For performance evaluation, following standard practices in existing methods, we adopted standard Re-ID metrics, including Cumulative Matching Characteristic (CMC) curves and mean Average Precision (mAP).

\subsection{Benchmark Methods}

\par To validate the effectiveness of our method, we follow the comparison protocols adopted in the original papers of each dataset. We compare our method with the following SOTA methods: CM-NAS \cite{fu2021cm}, LbA \cite{park2021learning}, Hc-Tri \cite{liu2020parameter}, AGW \cite{ye2021deep}, DEEN \cite{zhang2023diverse}, MCJA \cite{liang2024bridging}, SOLIDER \cite{chen2023beyond}, ViT-base \cite{dosovitskiy2020image}, DeiT-base \cite{touvron2021training}, TransReID \cite{he2021transreid}, VersReID \cite{zheng2024versatile}, TransOSS \cite{wang2025cross}, DDAG \cite{ye2020dynamic}, CAJ \cite{ye2021channel}, MMD \cite{jambigi2021mmd}, and DART \cite{yang2022learning}.

\subsection{Comparison With SOTA Methods}

\par We compare our proposed method with existing SOTA methods on two cross-modality ship Re-ID datasets: HOSS-ReID and CMShipReID. Our method achieves SOTA performance in most settings with a minimal number of trainable parameters, demonstrating the effectiveness of our approach.

\begin{table*}[htbp]
	\centering
	\caption{Comparison with state-of-the-art methods on Hoss-ReID. The red numbers indicate the mAP gain over the baseline.}
	\label{tab:comparison_params}
	
	\definecolor{lighttext}{gray}{0.5}
	
	\scriptsize
	\renewcommand{\arraystretch}{1.1}
	\setlength{\tabcolsep}{1pt}

	\begin{tabular*}{\linewidth}{@{\extracolsep{\fill}} l c c c@{\extracolsep{6pt}}c@{\extracolsep{\fill}}ccc c@{\extracolsep{6pt}}c@{\extracolsep{\fill}}ccc c@{\extracolsep{6pt}}c@{\extracolsep{\fill}}ccc}
		\toprule
		\multirow{2}{*}{Backbone} & \multirow{2}{*}{Method} & Trainable & \multicolumn{5}{c}{All} & \multicolumn{5}{c}{Optical to SAR} & \multicolumn{5}{c}{SAR to Optical} \\
		\cmidrule(lr){4-8} \cmidrule(lr){9-13} \cmidrule(lr){14-18}
		&       & Params (M) & mAP   & & R1    & R5    & R10  & mAP   & & R1    & R5    & R10  & mAP   & & R1    & R5    & R10 \\
		\midrule
		\multirow{6}{*}{ResNet} 
		& CM-NAS &  -   & 30.7  &      & 46.0  & 54.6  & 57.4  & 8.2   &      & 1.5   & 10.8  & 21.5  & 7.6   &      & 4.5   & 11.9  & 19.4 \\
		& LbA   & 23.52 & 33.0  &      & 48.3  & 59.7  & 62.5  & 11.9  &      & 4.6   & 23.1  & 41.5  & 8.5   &      & 6.0   & 14.9  & 22.4 \\
		& Hc-Tri &   -   & 34.0  &      & 47.2  & 54.6  & 59.7  & 11.1  &      & 6.2   & 15.4  & 24.6  & 10.9  &      & 7.5   & 20.9  & 29.9 \\
		& AGW   & 33.72 & 43.6  &      & 57.4  & 64.2  & 68.8  & 17.2  &      & 7.7   & 29.2  & 38.5  & 21.1  &      & 14.9  & 34.3  & 46.3 \\
		& DEEN  &   -   & 43.8  &      & 58.5  & 64.2  & 66.5  & 31.3  &      & 21.5  & 44.6  & 60.0  & 27.4  &      & 22.4  & 40.3  & 53.7 \\
		& MCJA  & 26.40 & 47.1  &      & 59.1  & 67.9  & 73.0  & 18.6  &     & 10.8  & 27.7  & 38.5  & 19.7  &      & 14.9  & 28.3  & 43.3 \\
		\midrule
		\multirow{7}{*}{Transformer} 
		& SOLIDER &  -    & 38.2  &      & 50.6  & 63.1  & 69.9  & 23.1  &      & 12.3  & 38.5  & 52.3  & 14.6  &      & 10.4  & 16.4  & 31.3 \\
		& ViT-base & 86.00 & 43.0  &      & 56.2  & 64.8  & 69.9  & 21.5  &      & 12.3  & 33.8  & 55.4  & 17.9  &      & 10.4  & 25.4  & 32.8 \\
		& DeiT-base & 86.00 & 47.2  &      & 58.1  & 69.6  & 74.1  & 25.9  &      & 16.1  & 36.7  & 59.1  & 26.1  &      & 20.3  & 35.7  & 52.0 \\
		& TransReID & 86.00 & 48.1  &      & 60.8  & 69.3  & 73.9  & 27.3  &      & 18.5  & 40.0  & 58.5  & 20.9  &      & 11.9  & 34.3  & 43.3 \\
		& VersReID & 85.90 & 49.3  &      & 59.7  & 70.5  & 78.4  & 25.7  &      & 13.8  & 40.0  & 61.5  & 27.7  &      & 17.9  & 44.8  & 61.2 \\
		& TransOSS* & 86.00 & 49.4  &      & 61.9  & 71.0  & 78.4  & 30.2  &     & 16.9  & 49.2  & 63.1  & 29.1  &      & 20.9  & 49.3  & 64.2 \\
		& TransOSS & 86.00 & 57.4  &      & 65.9  & 79.5  & 85.8  & 48.9  &     & 33.8  & 67.7  & 80.0  & 38.7  &      & 29.9  & 59.7  & 71.6 \\
		\midrule
		\multirow{6}{*}{DINOv3} 
		
		& \color{lighttext} ViT-S & \color{lighttext} 0.14 & \color{lighttext} 34.4  
		&   %
		& \color{lighttext} 52.3  & \color{lighttext} 58.5  & \color{lighttext} 60.2  
		& \color{lighttext} 11.3  
		&   %
		& \color{lighttext} 4.6   & \color{lighttext} 21.5  & \color{lighttext} 26.2  
		& \color{lighttext} 5.8   
		&   %
		& \color{lighttext} 4.5   & \color{lighttext} 10.4  & \color{lighttext} 14.9 \\
		
		& DRI-S (Ours) & 1.54  & 57.9  
		& \scriptsize \textcolor{red}{+23.5} %
		& 67.0  & \textbf{80.1} & \textbf{87.5} 
		& 40.7  
		& \scriptsize \textcolor{red}{+29.4} %
		& 24.6  & 60.0  & 72.3  
		& \textbf{46.1}  
		& \scriptsize \textcolor{red}{+40.3} %
		& \textbf{43.3} & \textbf{67.2} & \textbf{77.6} \\

		& \color{lighttext} ViT-B & \color{lighttext} 0.28 & \color{lighttext} 33.1  
		&   %
		& \color{lighttext} 48.3  & \color{lighttext} 56.8  & \color{lighttext} 60.8  
		& \color{lighttext} 10.2  
		&   %
		& \color{lighttext} 6.2   & \color{lighttext} 16.9  & \color{lighttext} 21.5  
		& \color{lighttext} 4.7   
		&   %
		& \color{lighttext} 4.5   & \color{lighttext} 4.5   & \color{lighttext} 9.0 \\            
		
		& DRI-B (Ours) & 2.93  & 56.1  
		& \scriptsize \textcolor{red}{+23.0} 
		& 65.9  & 76.7  & 83.0  
		& 39.3  
		& \scriptsize \textcolor{red}{+29.1} 
		& 26.2  & 58.5  & 73.8  
		& 39.5  
		& \scriptsize \textcolor{red}{+34.8} 
		& 29.9  & 59.7  & 74.6 \\

		& \color{lighttext} ViT-L & \color{lighttext} 0.37 & \color{lighttext} 28.9  
		&   
		& \color{lighttext} 42.0  & \color{lighttext} 50.0  & \color{lighttext} 53.4  
		& \color{lighttext} 10.4  
		&   
		& \color{lighttext} 6.2   & \color{lighttext} 12.3  & \color{lighttext} 15.4  
		& \color{lighttext} 3.8   
		&   
		& \color{lighttext} 1.5   & \color{lighttext} 6.0   & \color{lighttext} 11.9 \\
		
		& DRI-L (Ours) & 7.05  & \textbf{60.5} 
		& \scriptsize \textcolor{red}{+31.6} 
		& \textbf{69.9} & 79.0  & 85.2  
		& \textbf{55.6} 
		& \scriptsize \textcolor{red}{+45.2} 
		& \textbf{43.1} & \textbf{76.9} & \textbf{86.2} 
		& 45.0  
		& \scriptsize \textcolor{red}{+41.2} 
		& 34.3  & 61.2  & 76.1 \\
		\bottomrule
	\end{tabular*}%
	
	{\footnotesize \vspace{4pt} \parbox{\linewidth}{The star * indicates TransOSS without pretraining on optical-SAR image pairs. ViT-S, ViT-B, and ViT-L respectively denote the baseline methods that use the frozen small, base, and large versions of the Vision Transformer as the backbone. $\Delta$ indicates the absolute improvement in mAP compared to the corresponding ViT baseline. “Optical to SAR” means that queries from the optical modality are searched only within the SAR modality gallery, and “SAR to Optical” has a similar meaning \cite{wang2025cross}.}}
	
\end{table*}

\subsubsection{Results on HOSS-ReID}

\par Table \ref{tab:comparison_params} presents the quantitative comparison on the HOSS-ReID dataset. We provide a comprehensive analysis from two perspectives: parameter efficiency and peak performance.

\par First, we highlight our DRI-S model. Despite being built on a smaller backbone (ViT-S) and utilizing only 1.54M trainable parameters—a mere 1.79\% of the 86.00M parameters used by the previous SOTA method TransOSS—DRI-S achieves remarkable results. In the comprehensive ``All'' setting, DRI-S surpasses TransOSS in both mAP (\textbf{57.9\%} vs. 57.4\%) and R1 (\textbf{67.0\%} vs. 65.9\%). More notably, in the challenging ``SAR to Optical'' setting, DRI-S demonstrates superior robustness, outperforming TransOSS by 7.4\% in mAP and 13.4\% in R1. While DRI-S slightly trails TransOSS in the ``Optical to SAR'' setting, DRI-S remains highly competitive with $\sim$50$\times$ fewer parameters strongly validates the efficacy of our feature-space injection strategy.

\par To explore the performance upper bound, we further compare our DRI-L (based on ViT-L) with TransOSS. It is crucial to note that although DRI-L uses a larger backbone, its trainable parameter count (7.05M) is still less than 10\% of TransOSS (86.00M), maintaining a significant advantage in training efficiency. As shown in Table \ref{tab:comparison_params}, DRI-L achieves comprehensive SOTA performance across all metrics. Particularly in the ``Optical to SAR'' setting where DRI-S was limited by capacity, DRI-L dominates TransOSS with a 6.7\% lead in mAP (\textbf{55.6\%} vs. 48.9\%) and a 9.3\% lead in R1 (\textbf{43.1\%} vs. 33.8\%).

\par Whether prioritized for extreme lightweight deployment (DRI-S) or maximum accuracy (DRI-L), our method consistently outperforms full fine-tuning baselines like TransOSS while requiring significantly fewer computational resources for training. This confirms that calibrating feature distributions is a more efficient path to cross-modality alignment than blindly optimizing high-dimensional weight spaces.

\begin{table*}[htbp]
	\centering
	\caption{Comparison with state-of-the-art methods on CMShipReID (Part I: TIR-VIS and NIR to VIS). The red numbers indicate the mAP gain over the baseline.}
	\label{tab:comparison_part1}
	
	\definecolor{lighttext}{gray}{0.5}
	\scriptsize
	\renewcommand{\arraystretch}{1.1}
	\setlength{\tabcolsep}{1pt}
	
	\begin{tabular*}{\linewidth}{@{\extracolsep{\fill}} l c c@{\hspace{-6pt}}c@{\extracolsep{\fill}}ccc c@{\hspace{-6pt}}c@{\extracolsep{\fill}}ccc c@{\hspace{-6pt}}c@{\extracolsep{\fill}}ccc}
		\toprule
		\multirow{2}{*}{Method} & Trainable & \multicolumn{5}{c}{TIR to VIS} & \multicolumn{5}{c}{VIS to TIR} & \multicolumn{5}{c}{NIR to VIS} \\
		\cmidrule(lr){3-7} \cmidrule(lr){8-12} \cmidrule(lr){13-17}
		& Params (M) & mAP & & R1 & R5 & R10 & mAP & & R1 & R5 & R10 & mAP & & R1 & R5 & R10 \\
		\midrule
		DDAG  & -  & 39.98 & & 25.26 & 55.77 & 72.54 & 42.70 & & 28.10 & 58.24 & 74.39 & 56.76 & & 43.37 & 72.74 & 86.02 \\
		CAJ   & -  & 45.44 & & 31.13 & 61.23 & 75.87 & 43.44 & & 29.58 & 58.16 & 72.50 & 67.60 & & 55.75 & 82.21 & 91.91 \\
		AGW   & 33.72  & 49.04 & & 33.79 & 67.39 & 80.61 & 46.54 & & 31.41 & 64.10 & 78.79 & 61.69 & & 49.44 & 76.52 & 86.90 \\
		LbA   & 23.52  & 48.95 & & 32.68 & 69.20 & 84.61 & 46.78 & & 29.71 & 68.03 & 83.69 & 64.31 & & 50.33 & 82.67 & 92.61 \\
		MMD   & -  & 51.94 & & 37.42 & 68.96 & 82.82 & 57.00 & & 41.53 & 76.07 & 88.45 & 66.73 & & 54.33 & 82.30 & 91.95 \\
		DART  & -  & 59.73 & & 45.30 & 77.30 & 87.94 & 58.35 & & 43.73 & 76.10 & 87.46 & 71.39 & & 59.24 & 87.52 & 95.58 \\
		DEEN  & -  & 61.00 & & 46.55 & 79.52 & 89.36 & 62.58 & & 47.47 & 81.79 & 92.07 & 68.52 & & 54.65 & 87.16 & 95.73 \\
		\midrule
		
		\color{lighttext} ViT-S & \color{lighttext} 0.04 & \color{lighttext} 31.1 
		& %
		& \color{lighttext} 56.5 & \color{lighttext} 77.4 & \color{lighttext} 83.1 
		& \color{lighttext} 30.6 
		& %
		& \color{lighttext} 58.9 & \color{lighttext} 72.5 & \color{lighttext} 78.1 
		& \color{lighttext} 16.1 
		& %
		& \color{lighttext} 29.6 & \color{lighttext} 54.6 & \color{lighttext} 67.4 \\
		
		DRI-S (Ours) & 1.44 & 67.1 
		& \scriptsize \textcolor{red}{+36.0} %
		& 95.6 & 98.1 & 99.5 
		& 66.6 
		& \scriptsize \textcolor{red}{+36.0} %
		& 91.3 & 97.0 & 98.6 
		& 42.6 
		& \scriptsize \textcolor{red}{+26.5} %
		& 76.6 & 92.7 & 96.0 \\
		
		\color{lighttext} ViT-B & \color{lighttext} 0.08 & \color{lighttext} 35.8 
		& %
		& \color{lighttext} 61.3 & \color{lighttext} 77.3 & \color{lighttext} 84.7 
		& \color{lighttext} 36.9 
		& %
		& \color{lighttext} 63.5 & \color{lighttext} 76.3 & \color{lighttext} 79.9 
		& \color{lighttext} 18.3 
		& %
		& \color{lighttext} 30.5 & \color{lighttext} 52.3 & \color{lighttext} 64.2 \\
		
		DRI-B (Ours) & 2.73 & 77.8 
		& \scriptsize \textcolor{red}{+42.0} %
		& \textbf{97.7} & 99.4 & \textbf{99.7} 
		& 78.1 
		& \scriptsize \textcolor{red}{+41.2} %
		& 95.5 & 98.2 & 99.2 
		& 60.6 
		& \scriptsize \textcolor{red}{+42.3} %
		& 85.8 & 95.4 & \textbf{97.7} \\
		
		\color{lighttext} ViT-L & \color{lighttext} 0.10 & \color{lighttext} 48.3 
		& %
		& \color{lighttext} 71.3 & \color{lighttext} 86.1 & \color{lighttext} 90.6 
		& \color{lighttext} 49.6 
		& %
		& \color{lighttext} 70.6 & \color{lighttext} 82.4 & \color{lighttext} 86.6 
		& \color{lighttext} 29.7 
		& %
		& \color{lighttext} 42.3 & \color{lighttext} 65.0 & \color{lighttext} 73.1 \\
		
		DRI-L (Ours) & 6.78 & \textbf{84.1} 
		& \scriptsize \textcolor{red}{+35.8} %
		& 97.6 & \textbf{99.5} & \textbf{99.7} 
		& \textbf{84.9} 
		& \scriptsize \textcolor{red}{+35.3} %
		& \textbf{97.0} & \textbf{99.2} & \textbf{99.5} 
		& \textbf{72.6} 
		& \scriptsize \textcolor{red}{+42.9} %
		& \textbf{87.7} & \textbf{96.4} & 97.4 \\
		\bottomrule
	\end{tabular*}%
	
	{\footnotesize \vspace{4pt} \parbox{\linewidth}{ ViT-S, ViT-B, and ViT-L respectively denote the baseline methods that use the frozen small, base, and large versions of the Vision Transformer as the backbone. The red numbers denote the absolute improvement in mAP compared to the corresponding ViT baseline. ``TIR to VIS'' means that queries from the TIR modality are searched only within the VIS modality gallery, and ``VIS to TIR'' , ``NIR to VIS'' has a similar meaning. }}
\end{table*}

\begin{table*}[htbp]
	\centering
	\caption{Comparison with state-of-the-art methods on CMShipReID (Part II: VIS to NIR and NIR-TIR). The red numbers indicate the mAP gain over the baseline.}
	\label{tab:comparison_part2}
	
	\definecolor{lighttext}{gray}{0.5}
	\scriptsize
	\renewcommand{\arraystretch}{1.1}
	\setlength{\tabcolsep}{1pt}
	
	\begin{tabular*}{\linewidth}{@{\extracolsep{\fill}} l c c@{\hspace{-6pt}}c@{\extracolsep{\fill}}ccc c@{\hspace{-6pt}}c@{\extracolsep{\fill}}ccc c@{\hspace{-6pt}}c@{\extracolsep{\fill}}ccc}
		\toprule
		\multirow{2}{*}{Method} & Trainable & \multicolumn{5}{c}{VIS to NIR} & \multicolumn{5}{c}{NIR to TIR} & \multicolumn{5}{c}{TIR to NIR} \\
		\cmidrule(lr){3-7} \cmidrule(lr){8-12} \cmidrule(lr){13-17}
		& Params (M) & mAP & & R1 & R5 & R10 & mAP & & R1 & R5 & R10 & mAP & & R1 & R5 & R10 \\
		\midrule
		DDAG  & -  & 57.34 & & 43.98 & 73.16 & 86.07 & 38.72 & & 23.68 & 54.51 & 71.90 & 39.69 & & 24.74 & 55.91 & 73.00 \\
		CAJ   & -  & 66.59 & & 54.65 & 81.13 & 91.07 & 35.02 & & 20.49 & 49.73 & 66.79 & 33.33 & & 18.97 & 47.54 & 66.14 \\
		AGW   & 33.72  & 62.03 & & 48.98 & 77.95 & 89.06 & 45.30 & & 29.85 & 63.16 & 78.41 & 44.92 & & 29.36 & 62.71 & 78.89 \\
		LbA   & 23.52  & 65.20 & & 50.87 & 83.71 & 93.42 & 41.94 & & 25.32 & 61.53 & 79.71 & 42.17 & & 25.74 & 61.46 & 79.52 \\
		MMD   & -  & 66.48 & & 53.71 & 82.70 & 92.31 & 50.80 & & 36.17 & 67.81 & 81.62 & 51.19 & & 37.22 & 68.86 & 83.32 \\
		DART  & -  & \textbf{72.18} & & 60.35 & 87.76 & 95.27 & 56.14 & & 40.36 & 74.40 & 86.51 & 57.73 & & 42.84 & 75.93 & 87.71 \\
		DEEN  & -  & 67.91 & & 54.65 & 85.12 & 93.94 & 61.47 & & 45.64 & 81.87 & 92.24 & 60.79 & & 45.30 & 80.74 & 92.21 \\
		\midrule
		
		\color{lighttext} ViT-S & \color{lighttext} 0.04 & \color{lighttext} 17.4 
		& %
		& \color{lighttext} 35.1 & \color{lighttext} 56.7 & \color{lighttext} 64.1 
		& \color{lighttext} 26.6 
		& %
		& \color{lighttext} 51.6 & \color{lighttext} 76.5 & \color{lighttext} 83.6 
		& \color{lighttext} 25.0 
		& %
		& \color{lighttext} 54.8 & \color{lighttext} 73.9 & \color{lighttext} 81.1 \\
		
		DRI-S (Ours) & 1.44 & 42.9 
		& \scriptsize \textcolor{red}{+25.5} %
		& 76.3 & 92.1 & 95.9 
		& 44.0 
		& \scriptsize \textcolor{red}{+17.4} %
		& 70.4 & 88.3 & 93.0 
		& 42.0 
		& \scriptsize \textcolor{red}{+17.0} %
		& 72.3 & 90.3 & 94.2 \\
		
		\color{lighttext} ViT-B & \color{lighttext} 0.08 & \color{lighttext} 20.7 
		& \scriptsize \textcolor{red}{+39.4} %
		& \color{lighttext} 43.3 & \color{lighttext} 59.9 & \color{lighttext} 65.5 
		& \color{lighttext} 29.3 
		& \scriptsize \textcolor{red}{+30.5} %
		& \color{lighttext} 54.1 & \color{lighttext} 76.1 & \color{lighttext} 82.6 
		& \color{lighttext} 26.4 
		& \scriptsize \textcolor{red}{+30.0} %
		& \color{lighttext} 55.5 & \color{lighttext} 76.3 & \color{lighttext} 81.5 \\
		
		DRI-B (Ours) & 2.73 & 60.1 
		& \scriptsize \textcolor{red}{+39.4} %
		& 84.0 & 94.5 & 97.6 
		& 59.8 
		& \scriptsize \textcolor{red}{+30.5} %
		& 82.2 & 93.1 & 96.1 
		& 56.4 
		& \scriptsize \textcolor{red}{+30.0} %
		& 81.1 & 94.5 & 96.9 \\
		
		\color{lighttext} ViT-L & \color{lighttext} 0.10 & \color{lighttext} 30.4 
		& %
		& \color{lighttext} 58.3 & \color{lighttext} 73.7 & \color{lighttext} 78.4 
		& \color{lighttext} 37.7 
		& %
		& \color{lighttext} 61.3 & \color{lighttext} 77.5 & \color{lighttext} 82.8 
		& \color{lighttext} 32.5 
		& %
		& \color{lighttext} 62.3 & \color{lighttext} 81.5 & \color{lighttext} 87.3 \\
		
		DRI-L (Ours) & 6.78 & 71.8 
		& \scriptsize \textcolor{red}{+41.4} %
		& \textbf{90.9} & \textbf{98.0} & \textbf{99.0} 
		& \textbf{67.5} 
		& \scriptsize \textcolor{red}{+29.8} %
		& \textbf{84.8} & \textbf{95.6} & \textbf{98.0} 
		& \textbf{64.3} 
		& \scriptsize \textcolor{red}{+31.8} %
		& \textbf{83.7} & \textbf{95.6} & \textbf{97.6} \\
		\bottomrule
	\end{tabular*}%
	
	{\footnotesize \vspace{4pt} \parbox{\linewidth}{ ViT-S, ViT-B, and ViT-L respectively denote the baseline methods that use the frozen small, base, and large versions of the Vision Transformer as the backbone. The red numbers denote the absolute improvement in mAP compared to the corresponding ViT baseline. “VIS to NIR” means that queries from the VIS modality are searched only within the NIR modality gallery, and “NIR to TIR” , “TIR to NIR” has a similar meaning. }}
\end{table*}

\subsubsection{Results on CMShipReID}

\par We perform a comprehensive evaluation on the CMShipReID dataset, which encompasses three distinct modalities: VIS, NIR, and TIR. This dataset presents a complex challenge due to the heterogeneous nature of the spectral data. As shown in Table \ref{tab:comparison_part1} and Table \ref{tab:comparison_part2}, our method achieves a sweeping victory across all protocols, demonstrating exceptional robustness.

\par Table \ref{tab:comparison_part1} reports the results on the TIR-VIS setting, which contains the largest modality gap (reflective vs. emissive). A striking observation is the unprecedented improvement in R1 accuracy. Our \textbf{DRI-S}, with a negligible \textbf{1.44M} trainable parameters, achieves a R1 score of \textbf{95.6\%} on ``TIR to VIS". Compared to the current SOTA method DEEN ($46.55\%$), this represents a staggering improvement of \textbf{49.05\%}. Similarly, for ``VIS to TIR", DRI-S outperforms DEEN by \textbf{43.83\%} in R1 ($91.3\%$ vs. $47.47\%$).

\par The massive gap between the moderate mAP gains ($\sim$6\%) and the explosive R1 gains ($\sim$49\%) suggests that previous methods (like DEEN) struggle to learn sufficiently discriminative features for precise instance matching, likely due to overfitting on modality-specific noise. In contrast, our DRI strategy effectively extracts structural, modality-invariant semantics (via the OE) and injects them into the frozen VFM. This preserves the VFM's powerful pre-trained discrimination capability while calibrating the distribution, ensuring that the top-1 retrieval is accurate even under extreme thermal-optical discrepancies.

\par For the NIR and VIS modalities (Table \ref{tab:comparison_part1} and \ref{tab:comparison_part2}), which are spectrally closer but still distinct, our method continues to dominate. Our \textbf{DRI-L} achieves SOTA performance with \textbf{6.78M} parameters. In the "NIR to VIS" setting, it surpasses the strong competitor DART by \textbf{1.21\%} in mAP and a remarkable \textbf{28.46\%} in R1 (\textbf{87.7\%} vs. $59.24\%$). In the ``VIS to NIR'' setting, while the mAP is comparable to DART, our R1 score sees a significant boost of \textbf{30.55\%} (\textbf{90.9\%} vs. $60.35\%$).

\par This indicates that while existing methods like DART can align global distributions (good mAP), they often compromise the fine-grained details necessary for precise ranking. Our feature-space optimization avoids altering the model weights, thereby preventing the distortion of fine-grained features and ensuring superior retrieval precision.

\par Table \ref{tab:comparison_part2} further validates our method on the less common NIR-TIR setting. DRI-L consistently achieves SOTA performance, outperforming DEEN by \textbf{6.03\%} in mAP and \textbf{39.16\%} in R1 for ``NIR to TIR''. Similarly, for ``TIR to NIR", DRI-L outperforms DEEN by \textbf{3.51\%} in mAP and \textbf{38.4\%} in R1.

\par The consistent performance gains across all three modality pairs (VIS-TIR, VIS-NIR, NIR-TIR) confirm the plug-and-play generalization ability of our DRI strategy. Regardless of the specific spectral gap, the learnable OE can dynamically capture the necessary domain shifts, proving that explicit feature calibration is a universal solution for CMS Re-ID.

\subsection{Ablation Study}
\par To thoroughly validate the effectiveness of our proposed DRI strategy, we conducted comprehensive ablation studies on the HOSS-ReID dataset. Furthermore, we investigated the architectural design of the OE, specifically analyzing the impact of depth and embedding dimension on performance and efficiency.

\begin{table}[htbp]
	\centering
	\caption{Ablation study of DRI on the HOSS-ReID dataset. ``SST'' denotes the ship size token. ``Full Fine-tuning'' means the backbone is trainable.}
	\label{tab:ablation_DRI}
	
	\resizebox{0.5\textwidth}{!}{
		\begin{tabular}{l c c c}
			\toprule
			Method & Trainable Params (M) & mAP & R1 \\
			\midrule
			ViT-S (Full Fine-tuning) & 28.82 & 52.8 & 62.5 \\
			
			\midrule
			
			ViT-S (Frozen Backbone) &  0.14 &  34.4 &  52.3 \\
			
			\quad + SST & 0.14 & 40.7 & 56.2 \\
			
			\quad + SST + LoRA \cite{hu2022lora} & 1.98 & 55.1 & 63.1 \\
			
			\quad + SST + Adapter \cite{houlsby2019parameter} & 2.51 & 54.1 & 64.8 \\
			
			\quad + SST + Mona \cite{yin20255} & 1.60 & 46.9 & 58.5 \\
			
			\midrule
			\rowcolor{gray!20}
			\quad + SST + DRI & 1.54 & \textbf{57.9} & \textbf{67.0} \\
			\bottomrule
		\end{tabular}
	}
\end{table}

\subsubsection{Effectiveness and Superiority of DRI}

\par In this section, we dismantle the proposed method to isolate the contribution of the DRI module. We also benchmark DRI against generic PEFT paradigms to highlight its task-specific advantages. To ensure a fair comparison, all generic PEFT methods were configured to maintain a trainable parameter budget comparable to that of DRI. Specifically, the settings were: LoRA ($rank=64$, $\alpha=64$, targeting all $q,k,v$ and projection layers) \cite{hu2022lora}, Adapter ($hiddenDim=128$) \cite{houlsby2019parameter}, and Mona \cite{yin20255} following its original configuration.

\par As reported in Table \ref{tab:ablation_DRI}, the baseline model (Frozen Backbone + SST) yields suboptimal results (34.4\% mAP). Integrating DRI significantly boosts performance, achieving a +23.5\% surge in mAP (\textbf{57.9\%}) and a +14.7\% increase in R1 (\textbf{67.0\%}). This confirms that explicitly injecting domain knowledge is crucial for adapting the frozen VFM to the cross-modality domain.

\par A counter-intuitive yet compelling finding is that DRI outperforms Full Fine-tuning. Despite utilizing only 1.54M parameters ($\sim$5.3\% of the 28.82M required for full fine-tuning), DRI achieves higher accuracy, surpassing the fully fine-tuned model by 5.1\% in mAP and 4.5\% in R1.

\par This phenomenon suggests that full fine-tuning on small-scale, data-scarce datasets like HOSS-ReID leads to overfitting and catastrophic forgetting of the VFM's generalizable features. In contrast, DRI preserves the robust pre-trained weights and only learns a lightweight offset, effectively acting as a regularizer that balances adaptation with generalization.

\par Compared to generic methods like LoRA (55.1\% mAP) and Adapter (54.1\% mAP), DRI achieves superior performance with fewer or comparable parameters. Specifically, DRI outperforms LoRA by 2.8\% in mAP and 3.9\% in R1 while using 0.44M fewer parameters.

\par Generic PEFTs like LoRA operate in the weight space, assuming that low-rank updates are sufficient. However, the significant modality gap in CMS Re-ID often violates this low-rank assumption. DRI operates in the feature space, directly calibrating activation distributions. This allows for more flexible and expressive adaptation to heterogeneous modalities, proving that feature injection is a more efficient path for cross-modality alignment than weight adaption.

\begin{table}[htbp]
	\centering
	\caption{Ablation study on the influence of depth and embedding dimension of OE. The default setting is Depth=2 and Dim=64 (Ours).}
	\label{tab:ablation_DE}
	\resizebox{0.48\textwidth}{!}{
		\begin{tabular}{c c c c c}
			\toprule
			Depth & Embed. Dim & Trainable Params (M) & mAP & R1 \\
			\midrule
			\rowcolor{gray!20}
			2 & 64 (Ours) & 1.54 & \textbf{57.9} & \textbf{67.0} \\
			\midrule
			\multicolumn{5}{l}{\textit{Impact of Depth}} \\
			1 & 64 & 1.49 & 55.3 & 64.8 \\
			4 & 64 & 1.64 & 56.1 & 65.9 \\
			6 & 64 & 1.74 & 56.8 & 64.2 \\
			\midrule
			\multicolumn{5}{l}{\textit{Impact of Embedding Dimension}} \\
			2 & 32 & 0.83 & 56.6 & 64.8 \\
			2 & 96 & 2.31 & 55.0 & 64.8 \\
			2 & 128 & 3.12 & 57.7 & 65.9 \\
			\bottomrule
		\end{tabular}
	}
\end{table}

\subsubsection{Architectural Design of OE}

\par We adopted a lightweight ViT architecture as the OE. To determine the optimal configuration, we investigate the sensitivity of the model to the OE's depth and embedding dimension. The results are summarized in Table \ref{tab:ablation_DE}.

\par We varied the depth of the OE from 2 to 6 layers. Results indicate that a shallow depth of 2 achieves the optimal balance between performance and efficiency, yielding the highest mAP of \textbf{57.9\%}. Increasing the depth to 4 or 6 does not result in significant performance gains but merely increases the trainable parameter count.

\par This finding suggests that the task of the OE—which is to learn a residual feature deviation ($\Delta \mathbf{x}$) rather than a full semantic representation—is relatively compact. A shallow encoder is sufficient to capture the necessary modality-specific statistics to bridge the domain gap. Furthermore, keeping the encoder shallow acts as a structural regularization, preventing the small module from overfitting to the limited training data while ensuring the injection remains parameter-efficient.

\par Similarly, fixing the depth at 2, we tested embedding dimensions of 64, 96, and 128. The compact dimension of 64 achieved the highest accuracy. Increasing the dimension to 128 resulted in comparable performance (57.7\% mAP) but doubled the parameter overhead.

\par This confirms that the domain representation vector $f_d$ can be highly compressed. A 64-dimensional vector effectively encodes the requisite modality and identity attributes to modulate the VFM. This compactness further validates our design philosophy: efficient cross-modality adaptation relies on precise calibration rather than high-dimensional redundancy.

\begin{figure}[t]
	\centering
	\includegraphics[width=0.98\linewidth]{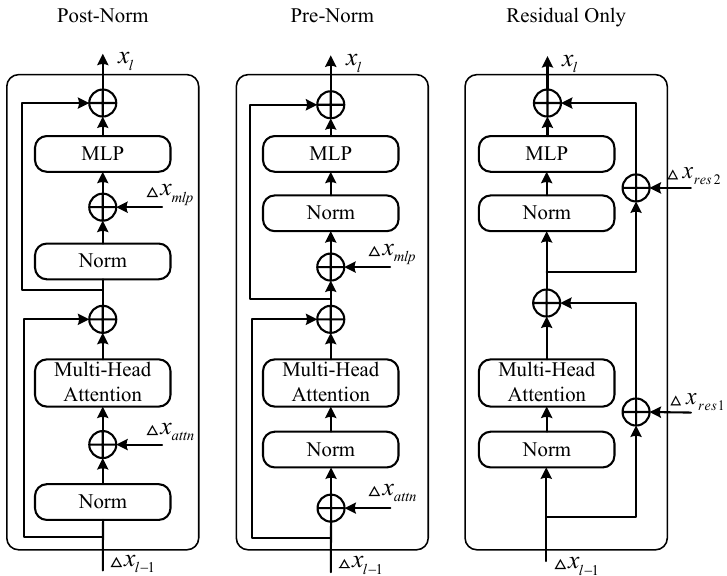} 
	\caption{\textbf{Schematic illustration of different injection positions investigated in the ablation study.} 
		\textbf{Post-Norm (Left):} Our proposed strategy where $\Delta \mathbf{x}$ is injected after Layer Normalization, serving as a direct calibration signal for the subsequent modules. 
		\textbf{Pre-Norm (Middle):} Injection is performed before Layer Normalization, where the learned deviation risks being suppressed by the normalization statistics. 
		\textbf{Residual Only (Right):} Injection is applied solely to the residual connections, shifting the output distribution without modulating the internal feature extraction process.}
	\label{fig:injection_variants}
\end{figure}

\begin{table}[t]
	\centering
	\caption{Ablation study on insertion positions. "Post-Norm" denotes injecting $\Delta \mathbf{x}$ after Layer Normalization (our default setting). "Pre-Norm" denotes injection before normalization.}
	\label{tab:deltax_position_ablation}
	\setlength{\tabcolsep}{3pt}
	\begin{tabular}{l l c c c}
		\toprule
		Target Module & Insertion Location & Trainable Params(M) & mAP & R1 \\
		\midrule
		Attention & Post-Norm & 0.94 & 53.4 & 61.4 \\
		MLP & Post-Norm & 0.94 & 56.4 & 66.5 \\
		\rowcolor{gray!20}
		Both (Ours) & Post-Norm & 1.54 & 57.9 & \textbf{67.0} \\
		\midrule
		Both & Pre-Norm & 1.54 & 35.3 & 45.5 \\
		Both & Residual Only & 1.54 & 52.4 & 59.7 \\
		Both & Residual + Pre-Norm & 2.74 & 37.1 & 47.7 \\
		Both & Residual + Post-Norm & 2.74 & \textbf{58.3} & 66.5 \\
		\bottomrule
	\end{tabular}
\end{table}

\subsubsection{Injection Positions of $\Delta \mathbf{x}$}

\par In the proposed DRI framework, we define the injection points at the inputs of the two core components within a ViT block: the MHSA and the MLP. To verify the optimality of this design, we conducted a comprehensive ablation study regarding \textit{where} to inject the learned domain deviations. As illustrated in Fig. \ref{fig:injection_variants}, we investigated three aspects: (1) target modules (Attention vs. MLP), (2) relative position to Layer Normalization (Pre-Norm vs. Post-Norm), and (3) alternative pathways (e.g., Residual connections). The results are summarized in Table \ref{tab:deltax_position_ablation}.

\par We first analyze the contribution of different modules. Injecting $\Delta \mathbf{x}$ solely into the Attention module yields an mAP of 53.4\%, while injecting into the MLP module achieves a higher mAP of 56.4\%.

\par This suggests that while calibrating spatial dependencies (Attention) is useful, adapting the channel-wise feature transformations (MLP) is more critical for cross-modality alignment. MLPs are often hypothesized to store knowledge and semantic patterns; thus, modulating them directly influences how domain-specific features are processed. Naturally, injecting into both modules yields the best performance (57.9\% mAP), indicating that spatial and channel-wise calibrations are complementary.

\par A striking observation is the drastic performance gap between Pre-Norm and Post-Norm injection strategies. As shown in Table \ref{tab:deltax_position_ablation}, injecting $\Delta \mathbf{x}$ before Layer Normalization (Pre-Norm) results in a catastrophic performance drop (35.3\% mAP), whereas injecting after Normalization (Post-Norm) achieves SOTA results (57.9\% mAP).

\par This phenomenon can be attributed to the mechanism of Layer Normalization, which re-centers and scales the input features. If $\Delta \mathbf{x}$ is injected before the Norm layer (Pre-Norm), the normalization statistics (mean and variance) will likely suppress or "wash out" the learned domain deviations, rendering the injection ineffective. In contrast, Post-Norm injection adds the deviation directly to the standardized features. This allows $\Delta \mathbf{x}$ to act as a robust "semantic prompt" or bias that directly guides the subsequent Attn/MLP computations, independent of the normalization scaling.

\par Finally, we explored injecting deviations into the residual connections ("Residual Only"). This strategy proves suboptimal (52.4\% mAP), likely because it only shifts the block output without intervening in the internal feature extraction process. While combining "Residual + Post-Norm" yields a marginal gain in mAP (58.3\% vs. 57.9\%), it causes a slight drop in Rank-1 and, more critically, nearly doubles the trainable parameters (2.74M vs. 1.54M).

\par Considering the trade-off between performance and parameter efficiency, the "Both + Post-Norm" strategy is the optimal choice, achieving the best alignment results with minimal computational overhead.

\subsubsection{Design of Modulator}

\begin{table}[t]
	\centering
	\caption{Ablation analysis of the Modulator design. "Zero" denotes initializing weights to zero to ensure an identity starting point. The results highlight the superiority of the simple Linear architecture combined with Zero initialization.}
	\label{tab:modulator_ablation_split}
	\setlength{\tabcolsep}{5pt}
	\begin{tabular}{l l c c c}
		\toprule
		Structure & Initialization & Trainable Params(M) & mAP & R1 \\
		\midrule
		Linear & Random & 1.54 & 47.7 & 59.7 \\
		MLP & Random & 1.64 & 49.7 & 60.8 \\
		MLP & Zero & 1.64 & 50.3 & 61.9 \\
		\rowcolor{gray!20}
		Linear & Zero (Ours) & 1.54 & \textbf{57.9} & \textbf{67.0} \\
		\bottomrule
	\end{tabular}
\end{table}

\begin{figure*}[t]
	\centering
	
	\includegraphics[width=0.98\textwidth]{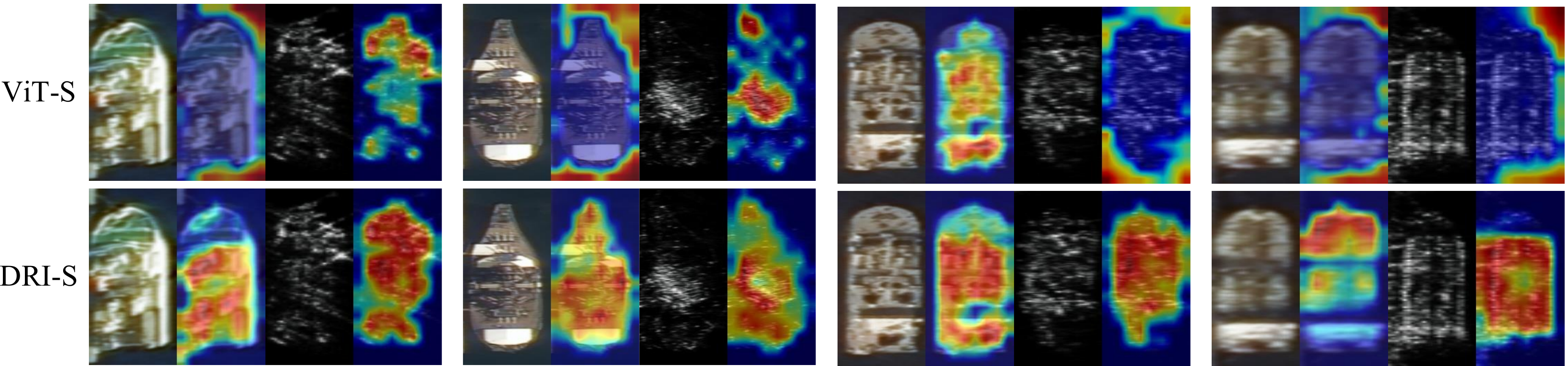}
	\caption{\textbf{Eigen-CAM visualization comparison on the HOSS-ReID dataset.} 
		The figure visualizes the feature map of the last ViT block.
		\textbf{Top Row (ViT-S):} The baseline model without adaptation.
		\textbf{Bottom Row (DRI-S):} Our proposed model equipped with Domain Representation Injection.
		From left to right, each row displays: the original RGB image, the RGB attention map, the original SAR image, and the SAR attention map.
		Observations show that the baseline (Top) is easily distracted by background clutter and SAR speckle noise, whereas DRI-S (Bottom) effectively suppresses noise and focuses on the structural topology of the ship hull, achieving precise semantic alignment across modalities.}
	\label{fig:visualization}
\end{figure*}

\par The Modulator serves as the bridge between the global domain representation $f_d$ and the layer-specific injection points. We explored two critical design aspects: the architectural complexity (Linear vs. MLP) and the initialization strategy (Random vs. Zero). The comparative results are presented in Table \ref{tab:modulator_ablation_split}.

\par Comparing the initialization strategies reveals a fundamental insight: initialization dictates the stability of adaptation. As shown in Table \ref{tab:modulator_ablation_split}, utilizing Random Initialization with a Linear structure results in poor performance (47.7\% mAP).

\par This is because randomly initialized modulators inject "noise" into the VFM at the beginning of training, significantly disturbing the pre-trained feature manifold. The model is forced to spend early epochs "unlearning" this noise. In contrast, Zero Initialization ensures that $\Delta \mathbf{x}$ starts at zero, effectively reverting the model to its original pre-trained state at step 0. This allows the DRI module to progressively learn useful domain deviations without inflicting an "initial shock" on the frozen backbone, leading to a substantial gain of +10.2\% in mAP (57.9\% vs. 47.7\%).

\par Regarding the architecture, a counter-intuitive observation is that the simple Linear projection significantly outperforms the more complex MLP. Even with Zero Initialization, the MLP variant lags behind the Linear variant by 7.6\% in mAP (50.3\% vs. 57.9\%).

\par While MLPs offer higher theoretical representational capacity, they also introduce optimization difficulties and a higher risk of overfitting, particularly in data-scarce scenarios like cross-modality ship Re-ID. We argue that the OE has already extracted high-level semantic attributes; therefore, the Modulator's role is merely to project these attributes into the channel dimensions of specific layers. A simple Linear transformation is sufficient for this projection and provides a smoother optimization landscape compared to an MLP. This adheres to the principle of parsimony: minimal complexity often yields maximum generalization in PEFT.

\subsubsection{Visualization and Interpretability}

\par To intuitively understand how DRI bridges the modality gap, we utilized Eigen-CAM \cite{muhammad2020eigen} to visualize the feature map of the last ViT block. Fig. \ref{fig:visualization} presents a qualitative comparison between the baseline ViT-S (Top Row) and our DRI-S (Bottom Row).

\par As observed in the top row, the frozen ViT-S exhibits significant vulnerability to domain shifts. In the RGB modality, the attention is loosely scattered, often focusing on the sea surface rather than the ship target. The failure is even more pronounced in the SAR modality. Due to the unique imaging mechanism of SAR, the background is filled with speckle noise and sea clutter. The baseline model, lacking domain-specific calibration, misinterprets these high-frequency noise patterns as discriminative features, resulting in chaotic activation maps that completely miss the ship's structure.

\par In stark contrast, the bottom row demonstrates the effectiveness of our DRI. With DRI, the attention mechanism is drastically corrected. In RGB images, the heatmap tightly wraps around the ship's body, filtering out the surrounding water. In SAR images, the improvement is critical: DRI effectively suppresses the inherent speckle noise. Instead of attending to random bright spots in the background, the model focuses sharply on the structural topology (e.g., the hull shape and orientation) of the ship.

\par This visualization reveals the core mechanism of our method: the learned domain representation $f_d$ acts as a structural filter. Since color and texture information is unreliable in SAR images, DRI forces the model to shift its attention from low-level texture (which is noisy in SAR) to high-level geometric shapes. This shape-aware alignment enables the model to extract consistent semantic features from heterogeneous modalities.

 \section{Conclusion} 
 
 \par In this paper, we addressed the challenges of significant modality discrepancies in CMS Re-ID. Departing from traditional weight-based PEFT paradigms, we proposed a novel DRI strategy that shifts the optimization perspective to the feature space. By designing a lightweight OE and adaptive Mod, our framework explicitly injects learnable domain deviations into a frozen VFM. This approach effectively bridges the modality gap while maximizing the preservation of pre-trained general knowledge. Extensive experiments on benchmark datasets demonstrate that our method achieves state-of-the-art performance with negligible training overhead (only 1.54M/6.78M trainable parameters). We hope this work inspires further exploration into feature-space adaptation for foundation models in cross-modal scenarios.

\bibliographystyle{IEEEtran}
\bibliography{ref}

\end{document}